\pdfoutput=1

\documentclass[11pt]{article}

\usepackage[]{acl}

\usepackage{times}
\usepackage{latexsym}
\usepackage{amsmath,amssymb}
\usepackage{graphicx}
\usepackage{hyperref}
\usepackage{cleveref}
\usepackage{comment}
\usepackage{subfig}
\usepackage{multicol,multirow,booktabs}
\usepackage[export]{adjustbox}
\usepackage{tikz}
\usepackage[round-mode=places]{siunitx}
\usetikzlibrary{angles, quotes}

\usepackage{ragged2e}
\usepackage[T1]{fontenc}

\usepackage[utf8]{inputenc}

\usepackage{microtype}

\title{Isotropy, Clusters, and Classifiers}

\author{Timothee Mickus$^{\heartsuit}$ \And Stig-Arne Grönroos$^{\heartsuit\spadesuit}$ \And
  Joseph Attieh$^{\heartsuit}$ \\[-1cm] \AND
  \null \\[-1cm]
  $^{\heartsuit}$ University of Helsinki, ~$^{\spadesuit}$ Silo.AI, ~Finland \\
  \texttt{firstname.lastname@helsinki.fi} \\
}

\begin{document}
\maketitle
\begin{abstract}

Whether embedding spaces use all their dimensions equally, i.e., whether they are isotropic, has been a recent subject of discussion.
Evidence has been accrued both for and against enforcing isotropy in embedding spaces.
In the present paper, we stress that isotropy imposes requirements on the embedding space that are not compatible with the presence of clusters---which also negatively impacts linear classification objectives.
We demonstrate this fact both empirically and mathematically and use it to shed light on previous results from the literature.

\end{abstract}

\section{Introduction}

Recently, there has been much discussion centered around whether vector representations used in NLP do and should use all dimensions equally.
This characteristic is known as isotropy: 
In an isotropic embedding model, every direction is equally probable, ensuring uniform data representation without directional bias.
At face value, such a characteristic would appear desirable: Naively, one could argue that an anisotropic embedding space would be overparametrized, since it can afford to use some dimensions inefficiently.

The debate surrounding isotropy was initially sparked by \citet{mu2018allbutthetop}, who highlighted that isotropic static representations fared better on common lexical semantics benchmarks, and \citet{ethayarajh-2019-contextual}, who stressed that contextual embeddings are anisotropic.
Since then, evidence has been accrued both for and against enforcing isotropy on embeddings.

In the present paper, we demonstrate that this conflicting evidence can be accounted for once we consider how isotropy relates to embedding space geometry.
Strict isotropy, as assessed by IsoScore \citep{rudman-etal-2022-isoscore}, requires the absence of clusters, and thereby also conflicts with linear classification objectives.  
This echoes previous empirical studies connecting isotropy and cluster structures \citep[a.o.]{ait-saada-nadif-2023-anisotropy}.
In the present paper, we formalize this connection mathematically in \Cref{sec:math}. 
We then empirically verify our mathematical approach in \Cref{sec:exp}, discuss how this relation sheds light on earlier works focusing on anisotropy in \Cref{sec:sota}, and conclude with directions for future work in \Cref{sec:ccl}.

\section{Some conflicting optimization objectives}
\label{sec:math}

We can show that isotropy---as assessed by IsoScore \citep{rudman-etal-2022-isoscore}---impose requirements that conflict with cluster structures---as assessed by silhouette scores \citep{ROUSSEEUW198753}---as well as linear classifier objectives.

\paragraph{Notations.} In what follows, let $\mathcal{D}$ be a multiset of points in a vector space, $\Omega$ a set of labels, and $\ell : \mathcal{D} \to \Omega$ a labeling function that associates a given data-point in $\mathcal{D}$ to the relevant label.
Without loss of generality, let us further assume that $\mathcal{D}$ is PCA-transformed.
Let us also define the following constructs for clarity of exposition:
\begin{align*}
    \mathcal{D}_\omega &= \left\{ \mathbf{d} ~:~ \ell\left(\mathbf{d}\right) = \omega \right\} \\
    \mathrm{sign}(\omega, \omega') &= 
        \begin{cases}
            -1 \qquad \mathrm{if} ~ \omega = \omega' \\
            +1 \hfill \mathrm{otherwise}
        \end{cases}
\end{align*}
Simply put, $\mathcal{D}_\omega$ is the subset of points in $\mathcal{D}$ with label $\omega$, whereas the $\mathrm{sign}$ function helps delineate terms that need to be maximized (inter-cluster) vs. terms that need to be minimized (intra-cluster).

\subsection{Silhouette objective for clustering}
We can consider whether the groups as defined by $\ell$ are in fact well delineated by the Euclidean distance, i.e., whether they form natural clusters.
This is something that can be assessed through silhouette scores, which involve a \emph{separation} and a \emph{cohesion} score for each data-point.
The cohesion score consists in computing the average distance between the data-point and other members of its group, whereas separation consists in computing the minimum cohesion score the data-point could have received with any other label than the one it was assigned to.
More formally, let:
\begin{align*}
    \mathrm{cost}(\mathbf{d}, \mathcal{S}) &= \frac{1}{|\mathcal{S}|} \sum\limits_{\mathbf{d'} \in \mathcal{S}} \sqrt{\sum_i \left( \mathbf{d}_i - \mathbf{d'}_i\right)^2} 
\end{align*}
then we can define the silhouette for one sample as
\begin{align*}%
    \mathrm{coh}(\mathbf{d}) &= \mathrm{cost}\left(\mathbf{d}, \mathcal{D}_{\ell(\mathbf{d})} \setminus \{\mathbf{d}\}\right)\\
    \mathrm{sep}(\mathbf{d}) &= \min\limits_{\omega' \in \Omega \setminus \{\ell(\mathbf{d})\}}  \mathrm{cost}\left(\mathbf{d}, \mathcal{D}_{\omega'}\right) \\
    \mathrm{silhouette}(\mathbf{d}) &= \frac{\mathrm{sep}(\mathbf{d}) - \mathrm{coh}(\mathbf{d})}{\max \{\mathrm{sep}(\mathbf{d}),\mathrm{coh}(\mathbf{d}) \}}
\end{align*}

Or in other words, the silhouette score is maximized when separation cost ($\mathrm{sep}$) is maximized and cohesion cost ($\mathrm{coh}$) is minimized.
Hence, to maximize the silhouette score across the whole dataset $\mathcal{D}$, one needs to (i) maximize all inter-cluster distances, and (ii) minimize all intra-cluster distances.

We can therefore define a maximization objective for the entire set $\mathcal{D}$:
\begin{equation*}
    \sum\limits_{\mathbf{d} \in \mathcal{D}} \sum\limits_{\mathbf{d'} \in \mathcal{D}} \mathrm{sign}(\ell(\mathbf{d}), \ell(\mathbf{d'})) \sqrt{\sum_i \left( \mathbf{d}_i - \mathbf{d'}_i\right)^2} 
\end{equation*}
which, due to the monotonicity of the square root in $\mathbb{R}^+$, will have the same optimal argument $\mathcal{D}^\ast$ as the simpler objective $\mathcal{O}_\mathrm{S}$
\begin{equation} \label{eq:silhouettes-obj}
    \mathcal{O}_\mathrm{S} = \sum\limits_{\mathbf{d} \in \mathcal{D}} \sum\limits_{\mathbf{d'} \in \mathcal{D}} \mathrm{sign}(\ell(\mathbf{d}), \ell(\mathbf{d'})) \sum_i \left( \mathbf{d}_i - \mathbf{d'}_i\right)^2
\end{equation}

\subsection{Incompatibility with IsoScore}
How does the objective in \eqref{eq:silhouettes-obj} conflict with isotropy requirements?
Assessments of isotropy such as IsoScore  generally rely on the variance vector. 
As we assume $\mathcal{D}$ to be PCA transformed, the covariance matrix is diagonalized, and we can obtain variance for each individual component through pairwise squared distances \citep{6260326}:
\begin{equation*}
\mathbb{V}(\mathcal{D})_i = \frac{1}{2|\mathcal{D}|^2} \sum\limits_{\mathbf{d} \in \mathcal{D}} \sum\limits_{\mathbf{d'} \in \mathcal{D}} \left( \mathbf{d}_i  - \mathbf{d'}_i \right)^2
\end{equation*}

In IsoScore, this variance vector is then normalized to the length of the $\vec{1}$ vector of all ones, before computing the distance between the two:
\begin{equation*}
\sqrt{\sum_i \left(\frac{\lVert\vec{1}\rVert_2}{\lVert\mathbb{V}(\mathcal{D})\rVert_2}\mathbb{V}(\mathcal{D})_i - 1\right)^2}
\end{equation*}
This distance is taken as an indicator of isotropy defect, i.e., isotropic spaces will minimize it.

\makeatletter\ifacl@finalcopy
\begin{figure}
    \centering
    \begin{tikzpicture}
\coordinate  (O)  at (0,0);
\coordinate  (A)  at (2,0);
\coordinate  (B)  at (0,2);
\coordinate  (C)  at (-2,0);
\coordinate  (D)  at (0,-2);

\begin{scope}
    \clip (-2.1,0) rectangle (2.1,2.1);
    \draw (0,0) circle(2cm);
\end{scope}
\coordinate[label=above:$\color{red}\mathbf{c}_a$]  (M) at (80:2);
\coordinate[label=above left:$\color{blue}\mathbf{c}_b$]  (N) at (155:2);
\coordinate[label=above right:$\mathbf{r}$]  (R) at (35:2);

\draw[thick]  (O) -- (R);
\draw[thick, red]    (O) -- (M)  (M) -- (R);
\draw[thick, blue]    (O) -- (N)  (N) -- (R);
\pic[draw, angle eccentricity=1.3, angle radius=4mm, red] {angle = R--O--M};
\pic[draw, angle eccentricity=1.3, angle radius=6mm, blue] {angle = R--O--N};
\end{tikzpicture}
    \caption{Relation between angle and chord.}
    \label{fig:angle-v-chord}
\end{figure}
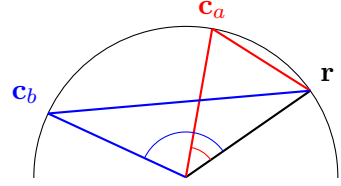
\fi

Given the normalization applied to the variance vector, the defect is computed as the distance between two points on a hyper-sphere. 
Hence it is conceptually simpler to think of this distance as an \emph{angle} measurement: 
Remark that as the cosine between $\mathbb{V}(\mathcal{D})$ and $\vec{1}$ increases, the isotropy defect decreases. \makeatletter\ifacl@finalcopy
A diagram illustrating this relation is provided in \Cref{fig:angle-v-chord}: 
For a given reference point $\mathbf{r}$ and two comparison points $\mathbf{c}_a$ and $\mathbf{c}_b$, we can observe that the shortest chord (from $\mathbf{r}$ to $\mathbf{c}_a$) also corresponds to the smallest angle.

More formally, let $\mathbf{\Tilde{v}} = \frac{\lVert\vec{1}\rVert_2}{\lVert\mathbb{V}(\mathcal{D})\rVert_2}\mathbb{V}(\mathcal{D})$ be the renormalized observed variance vector. 
We can note that both $\mathbf{\Tilde{v}}$ and the ideal variance vector $\vec{1}$ are points on the hyper-sphere centered at the origin and of radius $\lVert \vec{1} \rVert_2$.
As such, the defect is then equal to the distance between two points on a circle, i.e., the length of the chord between the renormalized observed variance vector and the ideal variance vector---which can be computed by simple trigonometry means, as $2\lVert \vec{1} \rVert_2 \sin\left(\alpha /2\right)$, with $\alpha$ the angle between $\mathbf{\Tilde{v}}$ and $\vec{1}$.
This can be converted to the more familiar cosine by applying a trigonometry identity (given that $0 \leq \alpha \leq \pi/4$):
\begin{align*}
    \lVert \mathbf{\Tilde{v}} - \vec{1} \rVert_2 &= 2\lVert \vec{1} \rVert_2\sqrt{1 - \cos^2 (\alpha / 2)} \\
    \frac{1}{4d} \lVert \mathbf{\Tilde{v}} - \vec{1} \rVert_2^2 - 1 &= -\cos^2 (\alpha / 2) 
\end{align*}
where $d$ is the dimension of the vectors in our point cloud.
Hence we can exactly relate the isotropic defect (squared) to the cosine (squared) of the angle between ideal and observed variance vectors. 

By monotonicity arguments, we can simplify this as follows:
\makeatother
To maximize isotropy, we have to maximize the objective $\mathcal{O}_\mathrm{I}$
\begin{align} \label{eq:isotropy-obj}
\mathcal{O}_\mathrm{I} 
    &= 
    \cos\left(\vec{1}, \mathbb{V}\left(\mathcal{D}\right)\right) \nonumber\\
    &\propto \sum\limits_{\mathbf{d} \in \mathcal{D}} \sum\limits_{\mathbf{d'} \in \mathcal{D}} \sum_i \left( \mathbf{d}_i  - \mathbf{d'}_i \right)^2
\end{align}
This intuitively makes sense: 
Ignoring vector norms, we have to maximize all distances between every pair of data-points  to ensure all dimensions are used equally, i.e., spread data-points out evenly on a hyper-sphere.
However, in the general case, it is not possible to maximize both the isotropy objective in \eqref{eq:isotropy-obj} and the silhouette score objective in \eqref{eq:silhouettes-obj}: 
Intra-cluster pairwise distances must be minimized for optimal silhouette scores, but must be maximized for optimal isotropy scores.
In fact, the two objectives can only be jointly maximized in the degenerate case where no two data-points in $\mathcal{D}$ are assigned the same label.\footnote{
    Hence some NLP applications and tasks need not be impeded by isotropy constrains, e.g., linear analogies that rely on vector offsets are \emph{a prima facie} compatible with isotropy.
}

\subsection{Relation to linear classifiers}

Informally, 
latent representations need to form clusters corresponding to the labels in order to optimize a linear classification objective.
Consider that in classification problems (i) any data-point $\mathbf{d}$ is to be associated with a particular label $\ell(\mathbf{d})=\omega_i$ and dissociated from other labels $\Omega  \setminus \{\ell(\mathbf{d})\}$, and (ii) association scores are computed using a dot product between the latent representation to be classified and the output projection matrix, where each column vector $\mathbf{c}^{\omega}$ corresponds to a different class label $\omega$.
As such, for any point $\mathbf{d}$ to be associated with its label $\ell(\mathbf{d})$, one has to maximize 
\begin{equation*} 
\resizebox{0.95\columnwidth}{!}{$%
    \langle \mathbf{d}, \mathbf{c}^{\ell(\mathbf{d})} \rangle = \frac{1}{2} \left(\lVert \mathbf{d} \rVert^2_2  + \lVert \mathbf{c}^{\ell(\mathbf{d})} \rVert^2_2 - \lVert \mathbf{d} - \mathbf{c}^{\ell(\mathbf{d})} \rVert^2_2 \right) 
$}%
\end{equation*}
In other words, one must either augment the norm of $\mathbf{d}$ or $\mathbf{c}^{\ell(\mathbf{d})}$, or minimize the distance between  $\mathbf{d}$ and $\mathbf{c}^{\ell(\mathbf{d})}$.
Note however that this does not factor in the other classes $\omega' \in \Omega \setminus \{\ell(\mathbf{d})\}$ from which $\mathbf{d}$ should be dissociated, i.e., where we must minimize the above quantity.
To account for the other classes, the global objective $\mathcal{O}_\mathrm{C}$ to maximize can  be defined as
\begin{equation}
\resizebox{0.95\columnwidth}{!}{$\begin{aligned}%
    \mathcal{O}_\mathrm{C} =& -\sum_{\mathbf{d} \in \mathcal{D}} \sum\limits_{\omega \in \Omega} \mathrm{sign}\left(\omega, \ell\left(\mathbf{d}\right)\right) \langle \mathbf{d}, \mathbf{c}^{\omega} \rangle \\
                =&  -\sum_{\mathbf{d} \in \mathcal{D}} \frac{|\Omega| - 2}{2}  \lVert \mathbf{d} \rVert^2_2  -\sum\limits_{\omega\in\Omega} \frac{|\mathcal{D}| - 2|\mathcal{D}_\omega|}{2} \lVert \mathbf{c}^{\omega} \rVert^2_2 \\
                 & + \frac{1}{2} \sum_{\mathbf{d} \in \mathcal{D}} \sum\limits_{\omega\in\Omega} \mathrm{sign}\left(\omega, \ell\left(\mathbf{d}\right)\right) \sum_i \left(\mathbf{d}_i - \mathbf{c}^{\omega}_i \right)^2  
\end{aligned}$}%
 \label{eq:classif-obj}
\end{equation}
\makeatletter\ifacl@finalcopy
where the weights $|\Omega| - 2$ and $|\mathcal{D}| - 2|\mathcal{D}_\omega|$ stem from counting how many other vectors a given data or class vector is associated with or dissociated from: we have one label to associate with any datapoint $\mathbf{d}$, and $|\Omega| - 1$ to dissociate it from; whereas a class vector $\mathbf{c}^\omega$ should be associated with the corresponding subset $\mathcal{D}_\omega$ and dissociated from the rest of the dataset (viz. $\mathcal{D} \setminus \mathcal{D}_\omega$).\footnote{
    The corresponding two sums can be understood as probabilistic priors over the data: 
    The objective entails that the norm of a class vector $\mathbf{c}^{\omega}$ should be proportional to the number of data-points with this label $\omega$, whereas one would expect a uniform distribution for vectors $\mathbf{d}$.  
    These terms cancel out for balanced, binary classification tasks.
}

\fi
Focusing on the last line of \Cref{eq:classif-obj}, we find that maximizing classification objectives entails minimizing the distance between a latent representation $\mathbf{d}$ and the vector for its label $\mathbf{c}^{\ell(d)}$, and maximizing its distance to all other class vectors.
It is reminiscent of the silhouette score in \Cref{eq:silhouettes-obj}: 
In particular any optimum for $\mathcal{O}_\mathrm{C}$ is an optimum for $\mathcal{O}_\mathrm{S}$, since it entails $\mathcal{D}^\ast$ such that 
\begin{equation}
\label{eq:optimum}
\forall \mathbf{d}, \mathbf{d'} \in \mathcal{D}^\ast \quad \ell(\mathbf{d}) = \ell(\mathbf{d'}) \iff \mathbf{d} = \mathbf{d'}    
\end{equation}
Informally: The cluster associated with a label should collapse to a single point.
Therefore the isotropic objective $\mathcal{O}_\mathrm{I}$ in \Cref{eq:isotropy-obj} is equally incompatible with the learning objective $\mathcal{O}_\mathrm{C}$ of a linear classifier.

\paragraph{In summary,} (i) point clouds cannot both contain well-defined clusters and be isotropic; and (ii) linear classifiers should yield clustered and thereby anisotropic representations.

\section{Empirical confirmation}
\label{sec:exp}
To verify the validity of our demonstrations in \Cref{sec:math}, we can optimize a set of data-points for a classification task using a linear classifier: 
We should observe an increase in silhouette scores, and a decrease in IsoScore. \makeatletter\ifacl@finalcopy
Note that we are therefore evaluating the behavior of parameters as they are optimized; i.e., we do not intend to test whether silhouettes and IsoScore behave as expected on held-out data. 
This both allows us to precisely test the argument laid out in \Cref{sec:math} and cuts down computational costs significantly.
\fi

\subsection{Methodology}

We consider four setups: 
(i) optimizing SBERT sentence embeddings \citep{reimers-gurevych-2019-sentence}\footnote{\texttt{all-MiniLM-L6-v2}} on the binary polarity dataset of \citet{pang-lee-2004-sentimental}; 
(ii) optimizing paired SBERT embeddings\footnotemark[\value{footnote}] on the validation split of SNLI \citep{bowman-etal-2015-large};
(iii) optimizing word2vec embeddings\footnote{\url{http://vectors.nlpl.eu/repository/}, model 222, trained on an English Wikipedia dump of November 2021.} on POS-tagging multi-label classification using the English CoDWoE dataset \citep{mickus-etal-2022-semeval};
and (iv) optimizing word2vec embeddings\footnotemark[\value{footnote}] for WordNet supersenses multi-label classification (\citealp{fellbaum-1998-wordnet}; pre-processed by \citealp{bylinina-etal-2023-leverage}).
All these datasets and models are in English and CC-BY or CC-BY-SA.\footnote{ 
    Our use is consistent with the intended use of these resources.
    We trust the original creators of these resources that they contain no personally identifying data.
}
Relevant information is available in \Cref{tab:sizes}; remark we do not split the data as we are interested on optimization behavior.
We also replicate and extend these experiments on GLUE in \Cref{adx:glue}.

\begin{table}[!t]
    \centering
    \resizebox{\linewidth}{!}{
    \begin{tabular}{l r r}
        \toprule
        \textbf{Dataset} & \textbf{N. items} & \textbf{N. params.} \\
        \midrule
         \citet{pang-lee-2004-sentimental} & \multirow{2}{*}{10~662} & \multirow{2}{*}{4~094~976} \\
         through \texttt{nltk} \citep{bird-loper-2004-nltk} \\
        \midrule
        \citet{bowman-etal-2015-large} & \multirow{2}{*}{9~842} & \multirow{2}{*}{4~987~395} \\
         from \href{https://nlp.stanford.edu/projects/snli/}{\tt nlp.stanford.edu} \\
         \bottomrule
        \citet{mickus-etal-2022-semeval} & \multirow{2}{*}{11~462} & \multirow{2}{*}{4~341~004} \\
         from \href{https://codwoe.atilf.fr/}{\tt codwoe.atilf.fr} \\
         \bottomrule
        \citet{fellbaum-1998-wordnet} & \multirow{2}{*}{2~275} & \multirow{2}{*}{690~326} \\
         from \href{https://github.com/altsoph/modality_shifts/tree/main}{\tt github.com/altsoph} \\
         \bottomrule
    \end{tabular}
    }
    \caption{Dataset vs. number of datapoints (N. items) and corresponding number of trainable parameters (N. params.).}
    \label{tab:sizes}
\end{table}

For (i) and (ii), we directly optimize the output embeddings of the SBERT model rather than update the parameters of the SBERT model. 
In all cases, we compute gradients for the entire dataset, and compute silhouette scores with respect to the target labels and IsoScore over 1000 updates.
In multi-label cases (iii) and (iv), we consider distinct label vectors as distinct target assignments when computing silhouette scores.
Models are trained using the Adam algorithm \citep{Kingma2014AdamAM};\footnote{
    Learning rate of $0.001$, $\beta$ of $(0.9, 0.999)$.
}
in cases (i) and (ii) we optimize cross-entropy, in cases (iii) and (iv), binary cross-entropy per label.
Remark that setups (ii), (iii) and (iv) subtly depart from the strict requirements laid out in \Cref{sec:math}.

Training per model requires between 10 minutes and 1 hour on an RTX3080 GPU; much of which is in fact devoted to CPU computations for IsoScore and silhouette scores values.
Hyperparameters listed correspond to default PyTorch values \citep{10.5555/3454287.3455008}, no hyperparameter search was carried out.
IsoScore is computed with the pip package \texttt{IsoScore} \citep{rudman-etal-2022-isoscore} on unpaired embeddings, silhouette scores with \texttt{scikit-learn} \citep{scikit-learn}.

\begin{figure}[ht]
    \centering
    \subfloat[Silhouette across training]{
        \includegraphics[max width=\columnwidth]{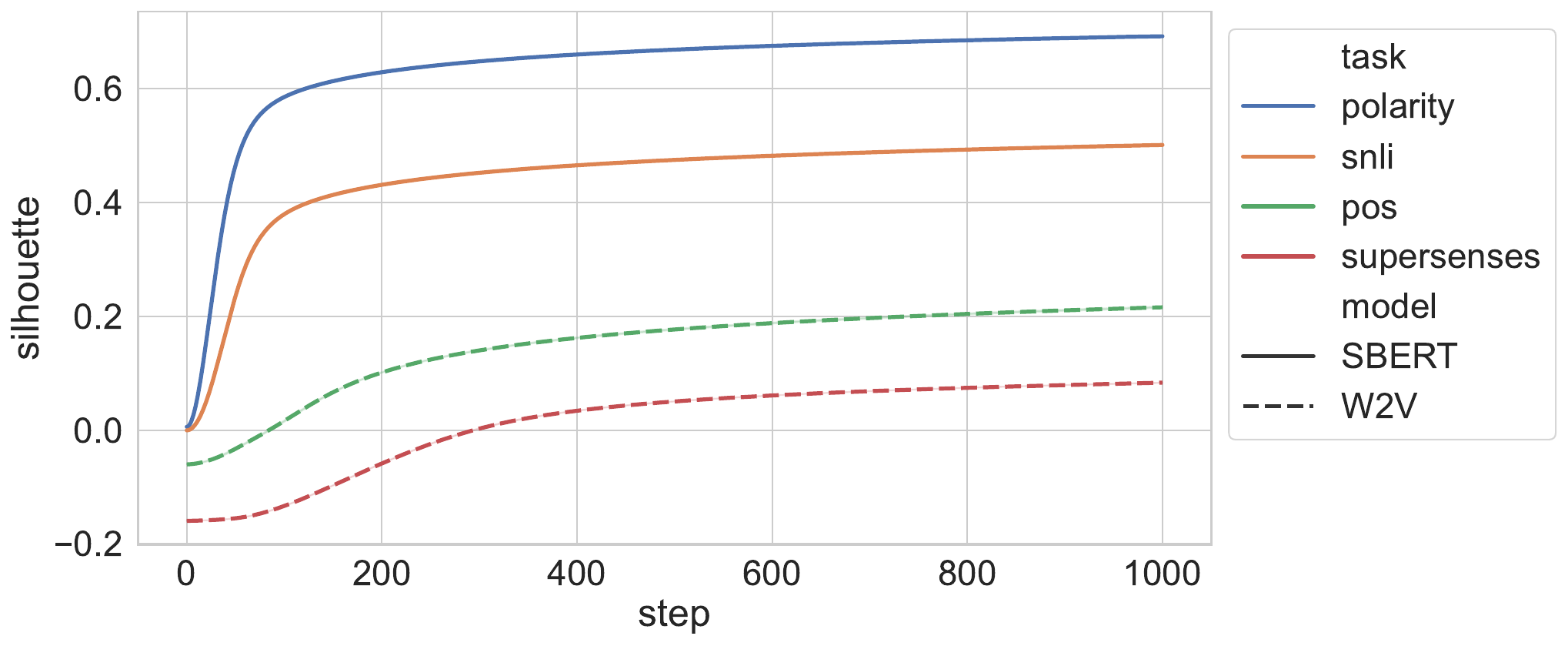}
    }

    \subfloat[Log-normalized IsoScore across training]{
        \includegraphics[max width=\columnwidth]{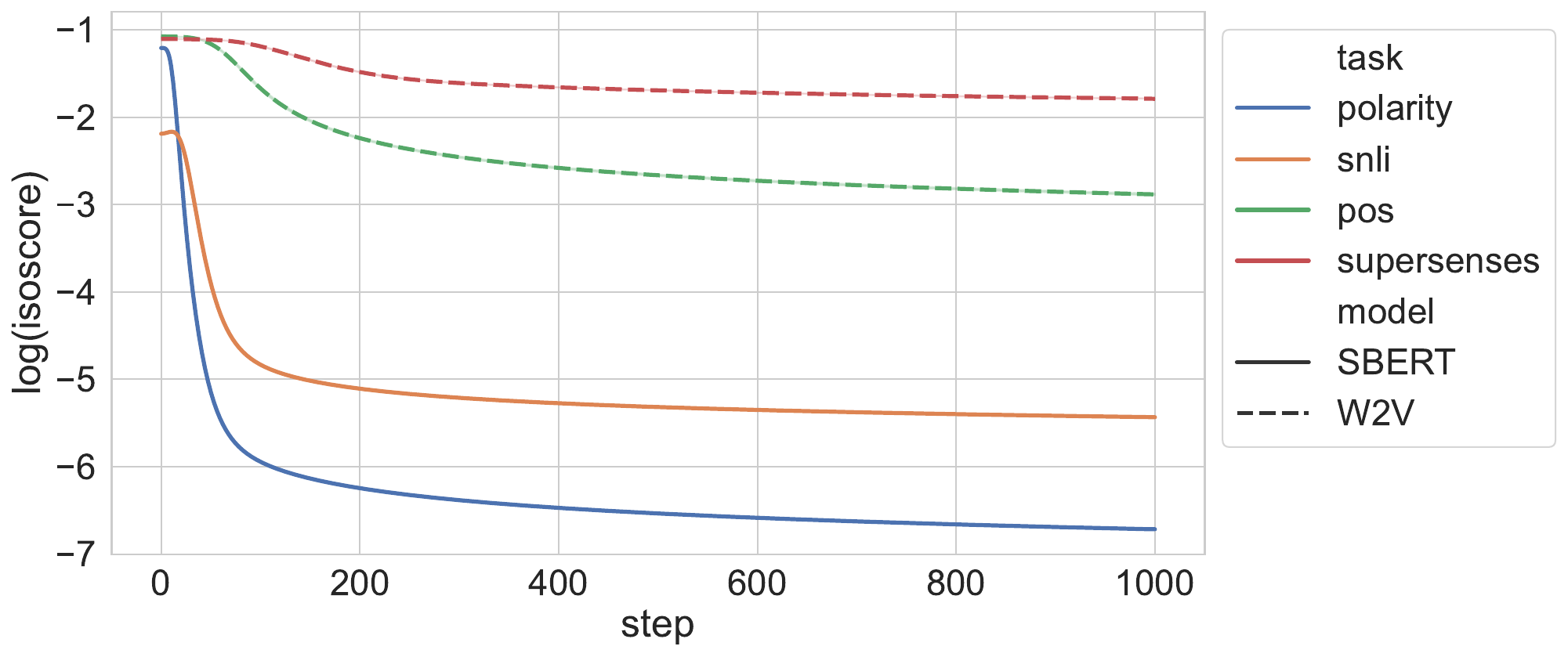}
    }

    \label{fig:res}
    \caption{Evolution of silhouette score and IsoScore across classification optimization (avg. of 5 runs).}
\end{figure}

\begin{figure}[ht]
    \centering
    \includegraphics[max width=0.75\columnwidth]{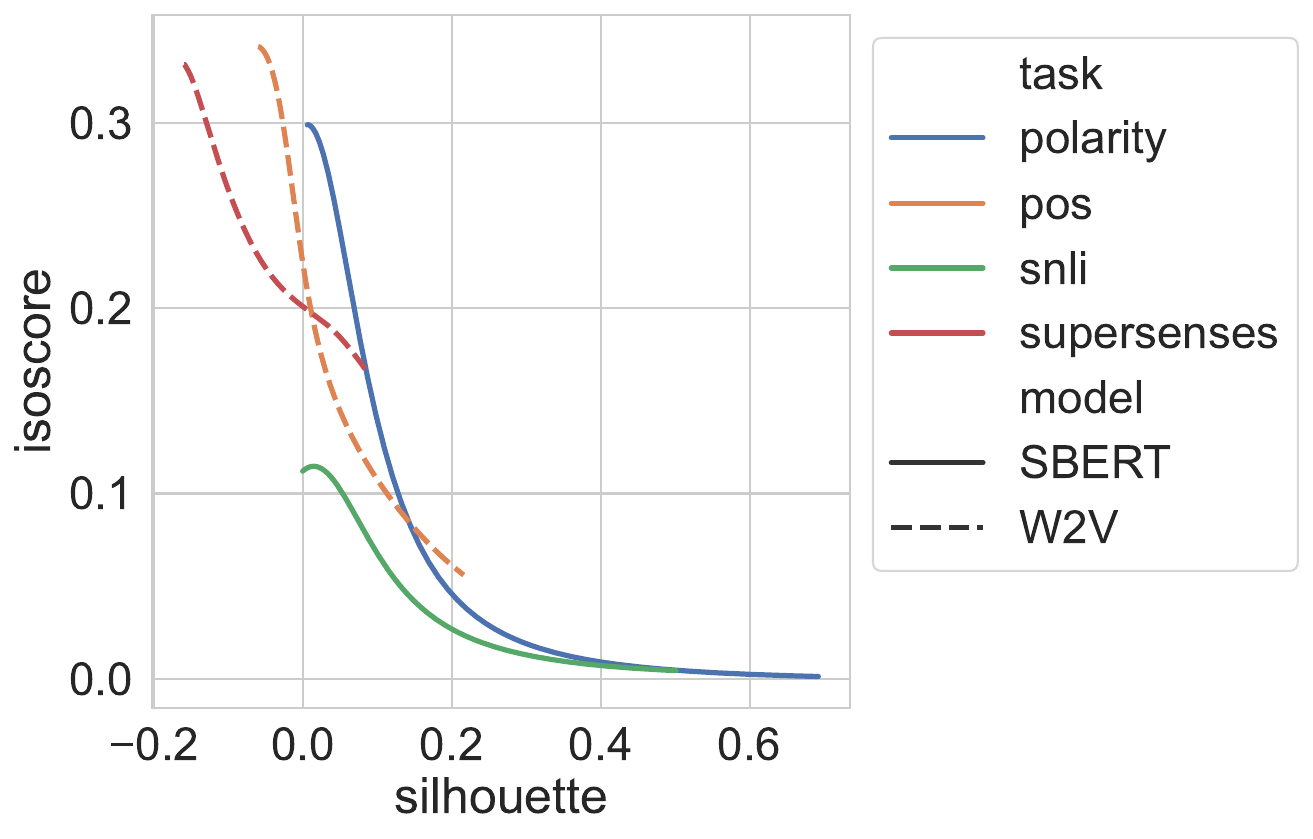}
    \caption{Relationship between silhouette scores and IsoScore (avg. of 5 runs).}
    \label{fig:rel}
\end{figure}

\subsection{Results}

Results of this empirical study are displayed in \Cref{fig:res}. 
Performances with five different random initialization reveal negligible standard deviations (maximum at any step $< 0.0054$, on average $< 0.0008$).
Our demonstration is validated: Across training to optimize classification tasks, the data-points become less isotropic and better clustered.
We can also see a monotonically decreasing relationship between IsoScore and silhouette scores, which is better exemplified in \Cref{fig:rel}:
We find correlations with Pearson's $r$ of $-0.808$ for the polarity task, $-0.878$ for SNLI, $-0.947$ for POS-tagging and $-0.978$ for supersense tagging; Spearman's $\rho$ are always below $-0.998$.

\paragraph{In summary,} we empirically confirm that isotropy requirements conflict with silhouette scores and linear classification objectives.

\section{Related works}
\label{sec:sota}
How does the connection between clusterability and isotropy that we outlined shed light on the growing literature on anisotropy?

While there is currently more evidence in favor of enforcing isotropy in embeddings, the case is not so clear cut that we can discard negative findings, and a vast majority of the positive evidence relies on improper techniques for quantifying isotropy \citep{rudman-etal-2022-isoscore}.
\citet{ethayarajh-2019-contextual} stressed that contextual embeddings are effective yet anisotropic.
\citet{ding-etal-2022-isotropy} provides experiments that advise against using isotropy calibration on transformers to enhance performance in specific tasks.
\citet{rudman2023stable} finds that anisotropy regularization in fine-tuning appears to be beneficial on a large array of tasks.
Lastly, \citet{rajaee-pilehvar-2021-cluster} find that the contrasts encoded in dominant dimensions can, at times, capture linguistic knowledge. 

On the other hand, the original study of \citet{mu2018allbutthetop} found that enforcing isotropy on static embeddings improved performances on semantic similarity, both at the word and sentence level, as well as word analogy.
Subsequently, a large section of the literature has focused on this handful of tasks \citep[e.g.,][]{liang-etal-2021-learning-to,timkey-van-schijndel-2021-bark}.
Isotropy was also found to be helpful beyond these similarity tasks: 
\citet{haemmerl-etal-2023-exploring} report that isotropic spaces perform much better on cross-lingual tasks, and \citet{jung-etal-2023-isotropic} stress its benefits for dense retrieval.

These are all applications that require graded ranking judgments, and therefore are generally hindered by the presence of clusters---such clusters would for instance introduce large discontinuities in cosine similarity scores.
To take \citet{haemmerl-etal-2023-exploring} as an example, note that language-specific clusters are antithetical to the success of cross-lingual transfer applications.
It stands to reason that isotropy can be found beneficial in such cases, although the exact experimental setup will necessarily dictate whether it is boon or bane: 
For instance \citet{rajaee-pilehvar-2021-fine-tuning} tested fine-tuning LLMs as Siamese networks to optimize performance on sentence-level similarity, and found enforcing isotropy to hurt performances---here, we can conjecture that learning to assign inputs to specific clusters is a viable solution in their case.

The literature has previously addressed the topic of isotropy and clustering. 
\citet{rajaee-pilehvar-2021-cluster} advocated for enhancing the isotropy on a cluster-level rather than on a global-level.  
\citet{cai2021isotropy} confirmed the presence of clusters in the embedding space with local isotropy properties. 
\citet{ait-saada-nadif-2023-anisotropy} investigated the correlation between isotropy and clustering tasks 
and found that fostering high anisotropy yields high-quality clustering representations. 
The study presented here provides a mathematical explanation for these empirical findings.


\section{Conclusion}
\label{sec:ccl}

We argued that isotropy and cluster structures are antithetical (\Cref{sec:math}), verified that this argument holds on real data (\Cref{sec:exp}), and used it to shed light on earlier results (\Cref{sec:sota}).
This result however opens novel and interesting directions of research: 
If anisotropic spaces implicitly entail cluster structures, then what is the structure we observe in our modern, highly anisotropic large language models?
Prior results suggest that this structure is in part linguistic in nature \citep{rajaee-pilehvar-2021-cluster}, but further confirmation is required.

Another topic we intend to pursue in future work concerns the relation between non-classification tasks and isotropy: 
Isotropy constraints have been found to be useful in problems that are not well modeled by linear classification, e.g. word analogy or sentence similarity.
Our present work does not yet offer a thorough theoretical explanation why. 
\makeatletter
\ifacl@finalcopy
\section*{Acknowledgments}
\noindent
{ 
\begin{minipage}{0.1\linewidth}
    \vspace{-10pt}
    \raisebox{-0.2\height}{\includegraphics[trim =32mm 55mm 30mm 5mm, clip, scale=0.18]{logos/erc.ai}} \\[0.25cm]
    \raisebox{-0.25\height}{\includegraphics[trim =0mm 5mm 5mm 2mm,clip,scale=0.075]{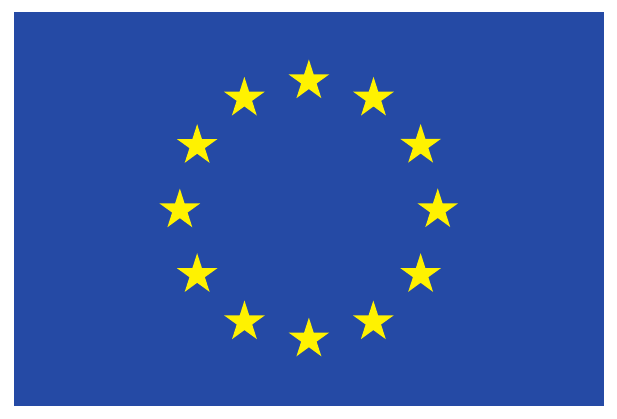}}
\end{minipage}
\hspace{0.01\linewidth}
\begin{minipage}{0.85\linewidth}
This work is part of the FoTran project, funded by the European Research Council (ERC) under the EU's Horizon 2020 research and innovation program (agreement \textnumero{}~771113). ~~We ~also ~thank ~the CSC-IT\vspace{0.3ex}
\end{minipage}
\begin{minipage}{\linewidth}
\noindent  Center for Science Ltd., for computational resources.
\end{minipage}
}
\fi
\makeatother

\section*{Limitations}
The present paper leaves a number of important problems open.

\makeatletter\ifacl@finalcopy

\paragraph{Idealized conditions.} Our discussion in \Cref{sec:math} points out optima that are incompatible, but says nothing of the behavior of models trained until convergence on held out data. In fact, enforcing isotropy could be argued to be a reasonable regularization strategy in that it would lead latent representations to not be tied to a specific classification structure.

Relatedly, a natural point of criticism to raise is whether our reasoning will hold for deep classifiers with non-linearities:
Most (if not all) modern deep-learning classification approaches rely on non-linear activation functions across multiple layers of computations. The present demonstration has indeed yet to be expanded to account for such more common cases.

Insofar neural architectures trained on classification objectives are concerned, we strongly conjecture their output embeddings would tend to be anisotropic. 
The anisotropy of inner representations appears to be a more delicate question: 
For Transformers, there has been extensive work showcasing that their structure is for the most part additive \citep{ferrando-etal-2022-towards,ferrando-etal-2022-measuring,modarressi-etal-2022-globenc,mickus-etal-2022-dissect,oh-schuler-2023-token,yang-etal-2023-local,mickus-vazquez-2023-bother}, and we therefore expect anisotropy to spread to bottom layers to some extent. 
For architectures based on warping random distributions such as normalizing flows \citep{Kobyzev_2021}, GANs \citep{NIPS2014_5ca3e9b1}, or diffusion models \citep{NEURIPS2020_4c5bcfec}, the fact that (part of) their input is random and isotropic likely limits how anisotropic their inner representations are.

\paragraph{Thoroughness of the mathematical framework.} The mathematical formalism is not thorough. 
For the sake of clarity and given page limitations, we do not include a formal demonstration 
that the linear classification optimum necessarily satisfies the clustering objective. Likewise, 
when discussing isotropy in \Cref{eq:isotropy-obj}, 
we ignore the cosine denominator.

\paragraph{Choice of objectives.}  Our focus on silhouette scores and linear classifier objectives may seem somewhat restrictive.
Our use of the silhouette score in the present derivation is motivated by two facts.
First, our interest is in how the point cloud will cluster along the provided labels---this rules out any external evaluation metric comparing predicted and gold label, such as ARI \citep{Hubert1985} or purity scores.
Second, we can also connect silhouette scores to a broader family of clustering metrics 
such as the Dunn index \citep{dunn-1974-well}, the Cali\'nski--Harabasz index \citep{calinski-harabasz-1974-dendrite} or the Davies--Bouldin index \citep{davies-bouldin-1979-cluster}.
Silhouette scores have the added benefit of not relying on centroids in their formulation, making their relation to the variance vector $\mathbb{V}(\mathcal{D})$ more immediate. 
We conjecture that these other criteria could be accounted for 
by means of triangular inequalities, as they imply the same optimum layout $\mathcal{D}^\ast$ as \Cref{eq:optimum}.

As for our focus on the linear classifier objective, we stress this objective is a straightforward default approach; but see \Cref{adx:triplet-loss} for a discussion of triplet loss within a similar framework as sketched here.

\fi

\bibliography{anthology,custom}
\bibliographystyle{acl_natbib}

\appendix

\makeatletter\ifacl@finalcopy

\section{Supplementary experiments on GLUE}
\label{adx:glue}

We reproduce experiments described in \Cref{sec:exp} on GLUE tasks \citep{wang-etal-2018-glue}.\footnote{From \href{https://huggingface.co/datasets/nyu-mll/glue}{\tt huggingface.co}.}
We train our models on the provided training sets---hence we only consider tasks for which there is a training set (all but \texttt{ax}) and that correspond to a classification problem (all but \texttt{stsb}, a regression task); we remove all datapoints where no label is provided.
Given our earlier results, we limit training to 250 updates; we directly update sentence-bert output embeddings by computing gradients for the entire training set all at once.
We compute IsoScore and silhouette scores after every update;
to alleviate computational costs, they are evaluated on random samples of $20,000$ items whenever the training set is larger than this (samples are performed separately for each update).
We test three different publicly available pretrained SBERT models: \texttt{all-mpnet-base-v2} (referred to as ``\texttt{mpnet}'' in what follows), \texttt{all-distilroberta-v1} (viz. ``\texttt{roberta}'') and \texttt{all-MiniLM-L6-v2} (viz. ``\texttt{miniLM}'').
Training details otherwise match those of \Cref{sec:exp}; see \Cref{tab:sizes-glue} for further information on the number of datapoints and parameter counts of all models considered.

\begin{table}[t]
    \centering
    \resizebox{\linewidth}{!}{
    \begin{tabular}{>{\tt}l r r r r }
        \toprule
        \multirow{2}{*}{\normalfont \textbf{Dataset}} & \multirow{2}{*}{\normalfont \textbf{N. items}} & \multicolumn{3}{c}{\textbf{N. params.}} \\
        && \tt miniLM & \tt mpnet & \tt roberta \\
        \midrule
        cola &   8~551 &   3~277~058 &   6~554~114 &   6~554~114 \\
        mnli & 392~702 & 199~380~483 & 398~760~963 & 398~760~963 \\
        mrpc &   3~668 &   2~709~506 &   5~419~010 &   5~419~010 \\
        qnli & 104~743 &  42~617~090 &  85~234~178 &  85~234~178 \\
        qqp  & 363~846 & 189~649~154 & 379~298~306 & 379~298~306 \\
        rte  &   2~490 &   1~738~370 &   3~476~738 &   3~476~738 \\
        sst2 &  67~349 &  25~720~322 &  51~440~642 &  51~440~642 \\
        wnli &     635 &     356~738 &     713~474 &     713~474 \\
        \bottomrule
    \end{tabular}
    }
    \caption{Supplementary experiments on GLUE: Dataset vs. number of datapoints (N. items) and corresponding number of trainable parameters (N. params.).}
    \label{tab:sizes-glue}
\end{table}

\begin{figure}[t]
    \centering
    \subfloat[Silhouette across training]{
        \includegraphics[max width=\columnwidth]{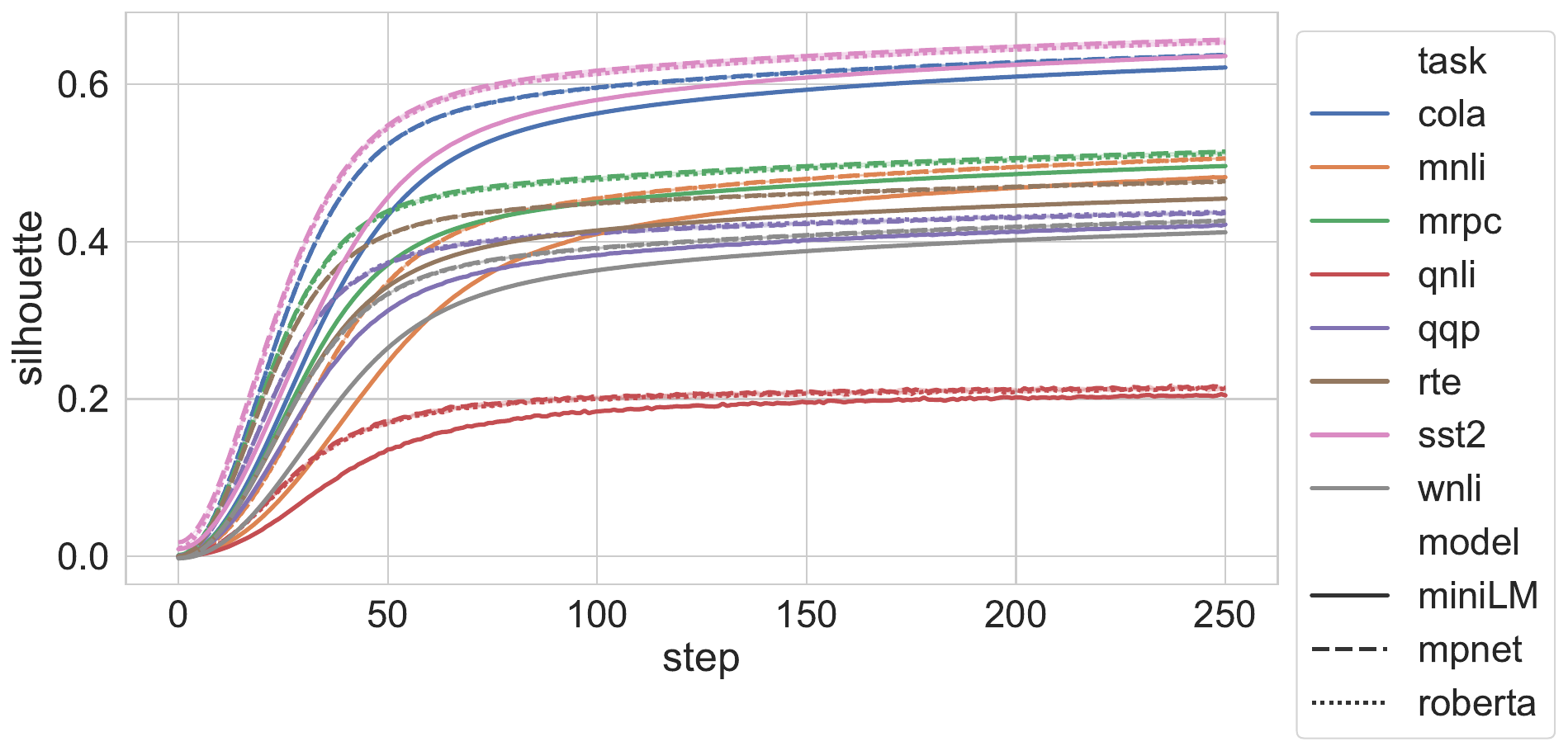}
    }

    \subfloat[Log-normalized IsoScore across training]{
        \includegraphics[max width=\columnwidth]{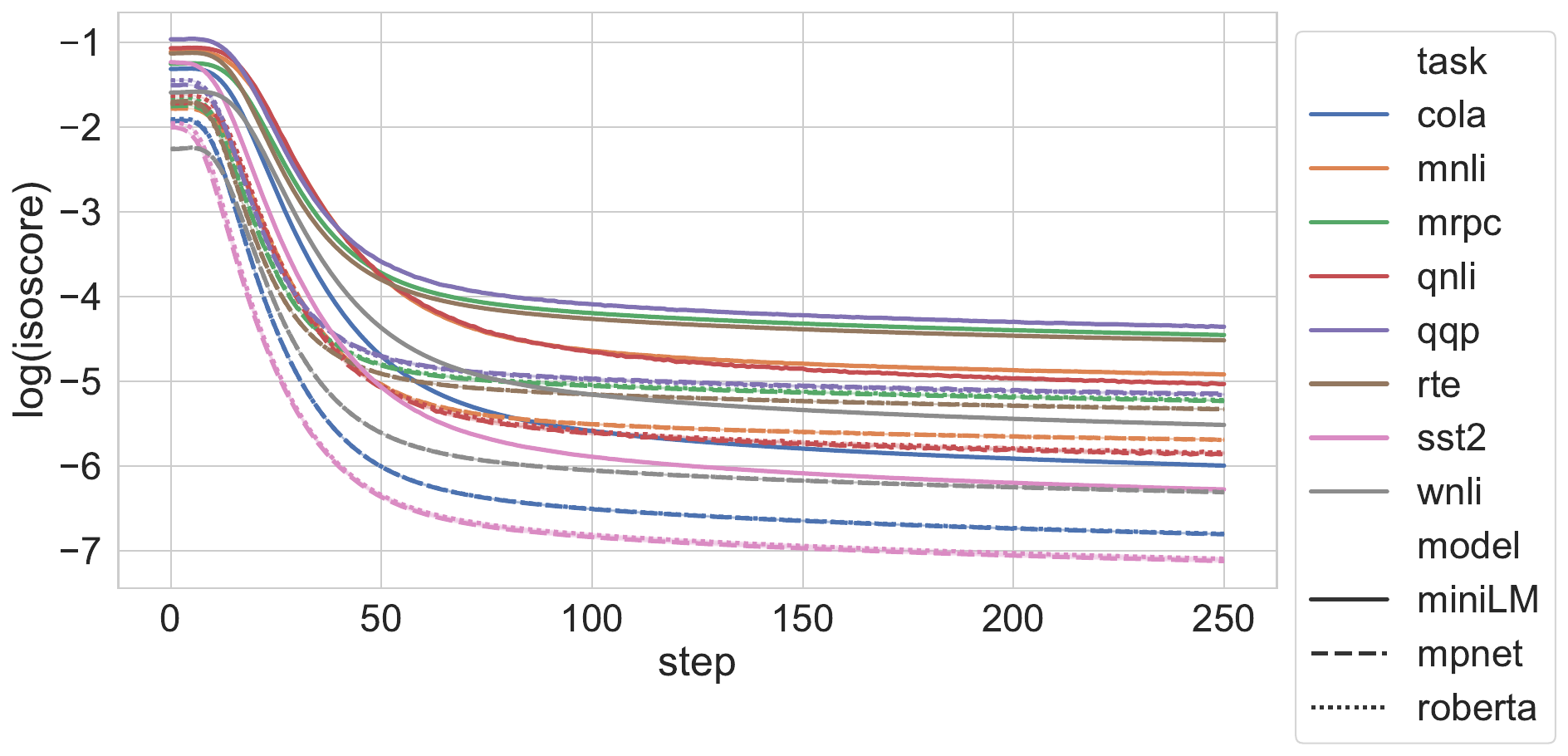}
    }

    \caption{Supplementary experiments on GLUE: Evolution of silhouette score and IsoScore across classification optimization (avg. of 5 runs).}
        \label{fig:exp-glue}
\end{figure}

\begin{table}[t]
    \centering
    \sisetup{round-precision=5}
    \begin{tabular}{l >{\tt}l S[table-format=2.5] S[table-format=2.5]}
        \toprule
             \multicolumn{2}{c}{\textbf{setup}} & $r$ & $\rho$ \\
        \midrule
           \multirow{8}{*}{\rotatebox{90}{\tt miniLM}} 
            & cola & -0.8829111906749023 & -0.9999637615057959 \\
 & mnli & -0.852172363973776 & -0.9993778413824672\\
 & mrpc & -0.9397301613400671 & -0.9966204648608807\\
 & qnli & -0.9118757105341755 & -0.9858763631325262\\
 & qqp & -0.9288981122707859 & -0.9966574199245969\\
 & rte & -0.9264835492305701 & -0.9998504276156709\\
 & sst2 & -0.8455079961211791 & -0.9999659758546953\\
 & wnli & -0.8968953620828243 & -0.9998689680279302\\
        \midrule
           \multirow{8}{*}{\rotatebox{90}{\tt mpnet}} 
  & cola & -0.8729893632432311 & -0.9999845860806568\\
 & mnli & -0.8445848713306897 & -0.9991953816319274\\
 & mrpc & -0.9245598135257743 & -0.9996991280678591\\
 & qnli & -0.905056680426905 & -0.9664952575496624\\
 & qqp & -0.9158306534991197 & -0.9950410271346989\\
 & rte & -0.9134839115853709 & -0.9998001456979615\\
 & sst2 & -0.8386415443466193 & -0.999948496324351\\
 & wnli & -0.8907657939361896 & -0.9999409684534762\\
\midrule
           \multirow{8}{*}{\rotatebox{90}{\tt roberta}} 
 & cola & -0.8713710099760137 & -0.9999853130664379\\
 & mnli & -0.8386460784340887 & -0.9991960828091486\\
 & mrpc & -0.9188334195884037 & -0.9984864900344953\\
 & qnli & -0.8991804077004847 & -0.9693823547296704\\
 & qqp & -0.9111499171354784 & -0.9942424446860518\\
 & rte & -0.915152188097229 & -0.999405234591852\\
 & sst2 & -0.8410313228652802 & -0.9999453212631838\\
 & wnli & -0.8902033477673262 & -0.9999115232538537\\
        \bottomrule
    \end{tabular}
    \caption{Supplementary experiments on GLUE: Correlations (Pearson's $r$ and Spearman's $\rho$) of IsoScore and silhouette scores in GLUE task}
    \label{tab:glue-corr}
\end{table}

Corresponding results are depicted in \Cref{fig:exp-glue}.
While there is some variation across models and GLUE tasks, all the setups considered display the same trend: Silhouette score increases and IsoScore decreases across training.
We can quantify this trend by computing correlation scores between IsoScore and silhouette scores.
Corresponding correlations are listed in \Cref{tab:glue-corr}: 
As is obvious, we find consistent and pronounced anti-correlations in all setups, with Pearson's $r$ always below $-0.838$ and Spearman's $\rho$ always below $-0.966$.
This further consolidates our earlier conclusions in \Cref{sec:exp}.

\section{Relation to triplet loss}
\label{adx:triplet-loss}
To underscore some of the limitations of our approach, we can highlight a connection with the triplet loss, which is often used to learn clusters.

It is defined for a triple of points $\mathbf{d}^a,\mathbf{d}^p, \mathbf{d}^n$ 
where $\ell(\mathbf{d}^a) = \ell(\mathbf{d}^p) \neq \ell(\mathbf{d}^n)$ as
\begin{equation*}
\resizebox{0.95\columnwidth}{!}{$\begin{aligned}%
    \mathcal{L}_{apn} &= \max\left( \lVert \mathbf{d}^a - \mathbf{d}^p\rVert_2 -  \lVert \mathbf{d}^a - \mathbf{d}^n\rVert_2, 0 \right) \\
     &= \max\left( \lVert \mathbf{d}^a - \mathbf{d}^p\rVert_2 , \lVert \mathbf{d}^a - \mathbf{d}^n\rVert_2 \right) -  \lVert \mathbf{d}^a - \mathbf{d}^n\rVert_2 \\
     &\geq \lVert \mathbf{d}^a - \mathbf{d}^p\rVert_2 -  \lVert \mathbf{d}^a - \mathbf{d}^n\rVert_2 \\
     &\hphantom{\geq\ }= \sum\limits_{\mathbf{d}_c \in \{\mathbf{d}^p, \mathbf{d}^n\}} - \mathrm{sign}\left(\ell\left(\mathbf{d}^a\right), \ell\left(\mathbf{d}_c\right)\right) \lVert \mathbf{d}^a - \mathbf{d}_c\rVert_2 
\end{aligned}$}%
\end{equation*}

The objective across the entire dataset $\mathcal{D}$ is thus:

\begin{equation}
\resizebox{0.95\columnwidth}{!}{$\begin{aligned}%
\mathcal{O}_\mathrm{T} &=
\sum\limits_{\omega \in  \Omega} \sum\limits_{\mathbf{d}^a \in \mathcal{D}_\omega} \sum\limits_{\mathbf{d}^p \in \mathcal{D}_\omega \setminus \{\mathbf{d}^a\}} \sum\limits_{\mathbf{d}^n \in \mathcal{D} \setminus \mathcal{D}_\omega}  - \mathcal{L}_{apn} \\
&\leq \sum\limits_{\omega \in  \Omega} \sum\limits_{\mathbf{d}^a \in \mathcal{D}_\omega} \sum\limits_{\mathbf{d}^p \in \mathcal{D}_\omega \setminus \{\mathbf{d}^a\}}  \sum\limits_{\mathbf{d}^n \in \mathcal{D} \setminus \mathcal{D}_\omega} \\
&\hphantom{\leq \sum\limits_{\omega \in  \Omega} } \sum\limits_{\mathbf{d}_c \in \{\mathbf{d}^p, \mathbf{d}^n\}} \mathrm{sign}\left(\ell\left(\mathbf{d}^a\right), \ell\left(\mathbf{d}^c\right)\right) \lVert \mathbf{d}^a - \mathbf{d}^c\rVert_2 \\
& \hphantom{\leq\ } = \sum\limits_{\mathbf{d} \in \mathcal{D}}  \sum\limits_{\mathbf{d}' \in \mathcal{D}} \mathrm{sign}_\mathrm{wgt}\left(\ell\left(\mathbf{d}\right), \ell\left(\mathbf{d}'\right)\right) \lVert \mathbf{d} - \mathbf{d}'\rVert_2
\end{aligned}$}%
\end{equation}
using a weighted variant of our original $\mathrm{sign}$ function:
$$ 
\mathrm{sign}_\mathrm{wgt}(\omega, \omega') = \begin{cases}	|\mathcal{D}_\omega | - |\mathcal{D}|  \qquad \mathrm{if} ~\omega = \omega' \\   	|\mathcal{D}_\omega| - 1 \hfill \mathrm{otherwise}	\end{cases}
$$

Remark that this is in fact an upper bound on both the silhouette objective as defined in \Cref{eq:silhouettes-obj} and the triplet objective $\mathcal{O}_\mathrm{T}$.
However, as they are to be maximized, the above does not entail that models trained with a triplet loss will necessarily develop anisotropic representations.



\fi

\end{document}